\documentclass[10pt,twocolumn,letterpaper]{article}

\usepackage{cvpr}
\usepackage{times}
\usepackage{epsfig}
\usepackage{graphicx}
\usepackage{amsmath}
\usepackage{amssymb}

\usepackage{bm}
\usepackage{tabularx} 
\newcolumntype{Y}{>{\centering\arraybackslash}X}

\usepackage{array,booktabs,arydshln,xcolor} 
\usepackage{amsthm}
\usepackage{epstopdf}
\usepackage{algorithm}
\usepackage{amsthm}
\usepackage{multirow}
\usepackage[algo2e]{algorithm2e}
\usepackage[outercaption]{sidecap}    
\usepackage{float}
\usepackage{wrapfig}
\usepackage{subcaption}
\usepackage{capt-of}
\usepackage{booktabs}
\usepackage{varwidth}

\usepackage[pagebackref=true,breaklinks=true,letterpaper=true,colorlinks,bookmarks=false]{hyperref}

\cvprfinalcopy 

\def\cvprPaperID{2871} 

\ifcvprfinal\fi
\begin{document}

\title{MDNet: A Semantically and Visually Interpretable \\ Medical Image Diagnosis Network}

\author{Zizhao Zhang, ~Yuanpu Xie, ~Fuyong Xing, ~Mason McGough, ~Lin Yang \\
	University of Florida\\
	{\tt\small zizhao@cise.ufl.edu}
}

\maketitle

\begin{abstract}
The inability to interpret the model prediction in semantically and visually meaningful ways is a well-known shortcoming of most existing computer-aided diagnosis methods. In this paper, we propose MDNet to establish a direct multimodal mapping between medical images and diagnostic reports that can read images, generate diagnostic reports, retrieve images by symptom descriptions, and visualize attention, to provide justifications of the network diagnosis process. MDNet includes an image model and a language model. The image model is proposed to enhance multi-scale feature ensembles and utilization efficiency. The language model, integrated with our improved attention mechanism, aims to read and explore discriminative image feature descriptions from reports to learn a direct mapping from sentence words to image pixels. The overall network is trained end-to-end by using our developed optimization strategy. Based on a pathology bladder cancer images and its diagnostic reports (BCIDR) dataset, we conduct sufficient experiments to demonstrate that MDNet outperforms comparative baselines. The proposed image model obtains state-of-the-art performance on two CIFAR datasets as well. 

\end{abstract}

\vspace{-.2cm} 
\section{Introduction}
\begin{figure}[t]
	\begin{center}
		\includegraphics[width=0.48\textwidth]{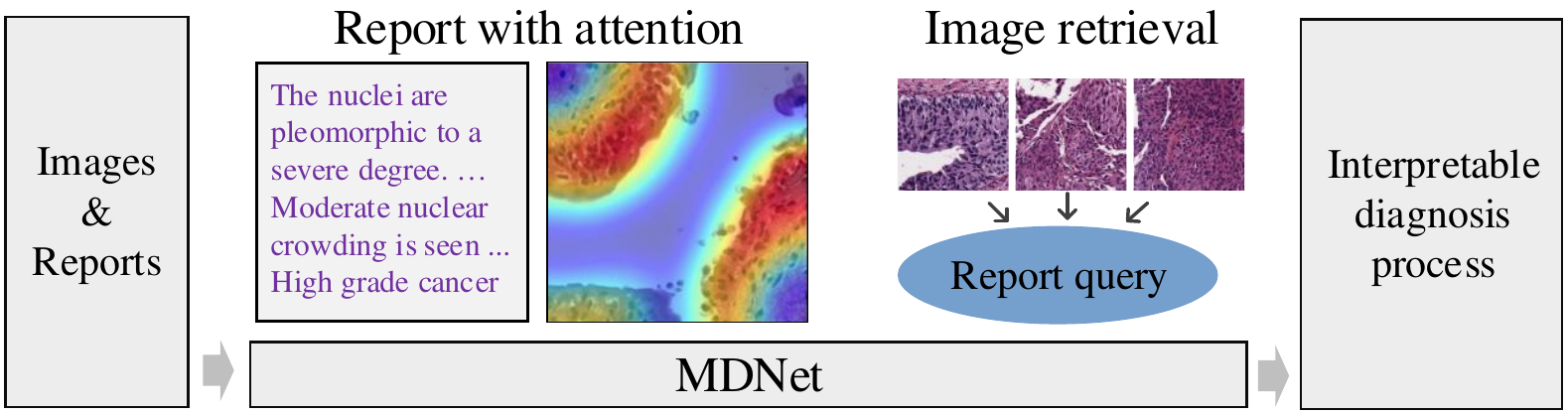}
		
	\end{center}
	\vspace{-.5cm}
	\caption{Overview of our medical image diagnosis network (MDNet) for interpretable diagnosis process. \label{fig:concept} } \vspace{-.4cm}
\end{figure}

\vspace{-.1cm}
In recent years, the rapid development of deep learning technologies has shown remarkable impact on the biomedical image domain. Conventional image analysis tasks, such as segmentation and detection \cite{cirecsan2013mitosis}, support quick knowledge discovery from medical metadata to help specialists' manual diagnosis and decision-making. Automatic decision-making tasks (e.g. diagnosis) are usually treated as standard classification problems. However, generic classification models are not an optimal solution for intelligent computer-aided diagnosis, because such models conceal the rationale for their conclusions, and therefore lack the interpretable justifications to support their decision-making process. It is rather difficult to investigate how well the model captures and understands the critical biomarker information.
A model that is able to visually and semantically interpret the underlying reasons that support its diagnosis results is significant and critical (Figure \ref{fig:concept}).
 
In clinical practice, medical specialists usually write diagnosis reports to record microscopic findings from images to diagnose and select treatment options. Teaching machine learning models to automatically imitate this process is a way to provide interpretability to machine learning models. Recently, image to language generation \cite{karpathy2015deep,mao2014deep,donahue2015long,vinyals2015show} and attention \cite{xu2015show} methods attract some research interests.


In this paper, we present a unified network, namely MDNet, that can read images, generate diagnostic reports, retrieve images by symptom descriptions, and visualize network attention, to provide justifications of the network diagnosis process. For evaluation, we have applied MDNet on a pathology bladder cancer image dataset with diagnostic reports (Section \ref{sec:dataset} introduces dataset details). 
In bladder pathology images, changes in the size and density of urothelial cell nuclei or thickening of the urothelial neoplasm of bladder tissue indicate carcinoma. Accurately describing these features facilitates the accurate diagnosis and is critical for the identification of early-stage bladder cancer. The accurate discrimination of those subtle appearance changes is challenging even for observers with extensive experience.  
To train MDNet, we address the problem of directly mining discriminative image feature information from reports and learn a direct multimodal mapping from report sentence words to image pixels. This problem is significant because discriminative image features to support diagnostic conclusion inference is ``latent'' in reports rather than offered by specific image/object labels. Effectively utilizing these semantic information in reports is necessary for effective image-language modeling. 
 
For image modeling based on convolutional neural networks (CNNs), we address the capability of the network to capture size-variant image features (such as mitosis depicted in pixels or cell polarity depicted in regions) for image representations. We analyze the weakness of the residual network (ResNet) \cite{he2015deep,he2016identity} from the ensemble learning aspect and propose \textit{ensemble-connection} to encourage multi-scale representation integration, which results in more efficient feature utilization according to our experiment results. 
For language modeling, we adopt Long Short-Term Memory (LSTM) networks \cite{vinyals2015show}, but focus on investigating the usage of LSTM to mine discriminative information from reports and compute effective gradients to guide the image model training. We develop an optimization approach to train the overall network end-to-end starting from scratch. 
We integrate the attention mechanism \cite{xu2015show}  in our language model and propose to enhance its visual feature alignment with sentence words to obtain sharper attention maps. 

To our knowledge, this is the first study to develop an interpretable attention-based model that can explicitly simulate the medical (pathology) image diagnosis process.
We perform sufficient experimental analysis with complementary evaluation metrics to demonstrate that MDNet can generate promising and reliable results, also outperforms well-known image captioning baselines \cite{karpathy2015deep} on the BCIDR dataset.  In addition, we validate the state-of-the-art performance of the proposed image model belonging to MDNet on two public CIFAR datasets \cite{krizhevsky2009learning}.

\section{Related Work}
\vspace{-.1cm}
\noindent \textbf{Image and language modeling:}  Joint image and language modeling enables the generation of semantic descriptions, which provides more intelligible predictions. Image captioning is one typical of application \cite{kiros2014multimodal}. Recent methods use recurrent neural networks (RNNs) to model natural language conditioned on image information modeled by CNNs \cite{karpathy2015deep,vinyals2015show,densecap,you2016image}. 
They typically employ pre-trained powerful CNN models, such as GoogLeNet \cite{simonyan2014very}, to provide image features. 
Semantic image features play a key role in accurate captioning \cite{mao2014deep,donahue2015long}.
Many methods focus on learning better alignment from natural language words to provided visual features, such as attention mechanisms \cite{xu2015show,you2016image,yang2015stacked}, multimodal RNN \cite{mao2014deep,karpathy2015deep,donahue2015long} and so on \cite{reed2016learning,yang2015stacked}. However, in the medical image domain, pre-trained universal CNN models are not available. A complete end-to-end trainable model for joint image-sentence modeling is an attractive open question, and it can facilitate multimodal knowledge sharing between the image and language models.

Image-sentence alignment also encourages visual explanations for network inner workings \cite{karpathy2014deep}. Hence, attention mechanisms become particularly necessary \cite{xu2015show}. We witness growing interests of its exploration to achieve the network interpretability \cite{zintgraf2016new,simonyan2013deep}. 
The full power of this field has vast potentials to renovate computer-aided medical diagnosis, but a dearth of related work exists. To date, \cite{shin2016learning} and \cite{kisilev2015semantic} deal with the problem of generating disease keywords for radiology images. 

\noindent \textbf{Skip-connection:} 
Based on the residual network (ResNet) \cite{he2015deep}, the new pre-act-ResNet \cite{he2016identity} introduces identity mapping skip-connection \cite{he2016identity} to address the network training difficulty. Identity mapping gradually becomes an acknowledged strategy to overcome the barrier of training very deep networks  \cite{he2016identity,huang2016deep,zagoruyko2016wide,huang2016densely}.  Besides, skip-connection encourages the integration of multi-scale representations for more efficient feature utilization \cite{long2015fully,bell2015inside,xie2015holistically}. 

\begin{figure*}[t]
	\begin{center}
		\includegraphics[width=0.98\textwidth]{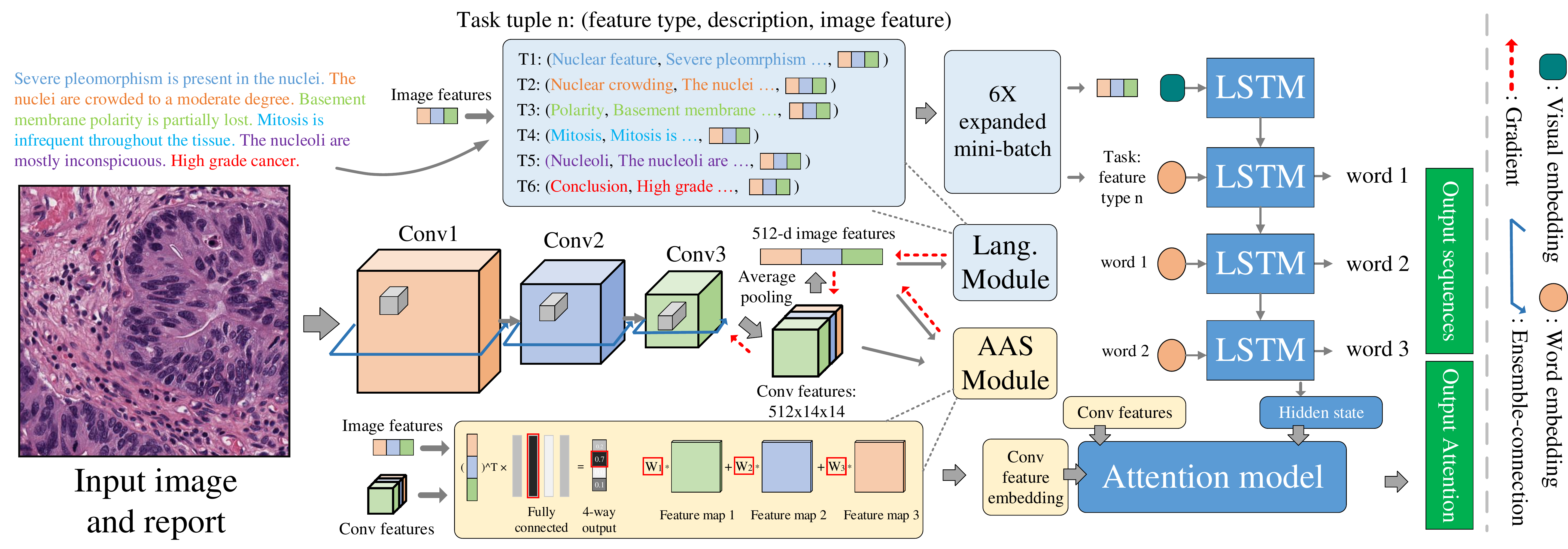}
		
	\end{center}
	\vspace{-.7cm}
	\caption{Overall illustration of MDNet. We use a bladder image with its diagnostic report as an example. The image model generates an image feature to pass to LSTM in the form of a task tuple and a Conv feature embedding (for the attention model) computed by the AAS module (defined in the method). LSTM executes prediction tasks according to the specified image feature type (best viewed in color).\vspace{-.5cm} \label{fig:archs} }
\end{figure*}

\section{Image model}
\label{sec:ecnet}
\vspace{-.1cm}
\subsection{Residual networks}
The identity mapping in the newest ResNet \cite{he2016identity} is a simple yet effective skip-connection to allow the unimpeded information flow inside the network \cite{srivastava2015highway}. Each skip-connected computation unit is called a residual block. In a ResNet with $L$ residual blocks, the forward output $y_{L}$ from the $l$-th residual block and the gradient of the loss $\mathcal{L}$ w.r.t its input $y_{l}$ is defined as
\vspace{-.6cm}
\begin{equation}
\label{eq:res_for}
y_{L} =  y_{l} + \sum_{m=l}^{L-1} \mathcal{F}_{m}(y_{m}), 
\end{equation}
\vspace{-.3cm}
\begin{equation}
\frac{\partial \mathcal{L}}{\partial y_l} = \frac{\partial \mathcal{L}}{\partial y_{L}} (1+ \frac{\partial}{\partial y_l}\sum_{m=l}^{L-1} \mathcal{F}_m(y_m)), \vspace{-.1cm}
\end{equation}
where $\mathcal{F}_{m}$ is composed by consecutive batch normalization \cite{ioffe2015batch}, rectified linear units (ReLU), and convolution. 
Thanks to the addition scheme, the gradient (i.e. $\frac{\partial \mathcal{L}}{\partial y_{L}}$) in backward can flow directly to preceding layers without passing through any convolutional layer. Since the weights of convolutional layers can scale gradients, this property alleviates the gradient vanishing effect when the depth of the network increases \cite{pascanu2013difficulty,he2016identity}. 

\subsection{Decouple ensemble network outputs}
\vspace{-.1cm}
One skip-connection in a residual block offers two information flow paths, so the total path increases exponentially as network goes deeper \cite{huang2016deep}. Recent work \cite{veit2016residual} shows that ResNet with $n$ residual blocks can be interpreted as the ensemble of $2^n$ relatively shallow networks.
It can be viewed that the exponential ensembles boost the network performance \cite{veit2016residual}. Consequently, this viewpoint reveals a weakness of ResNet by our probes into its classification module.


In ResNet and other related networks \cite{he2016identity,huang2016deep,lin2013network,szegedy2015going}, the classification module connecting convolutional layers includes a global average pooling layer and a fully connected layer. The two layers are mathematically defined as 
\vspace{-.1cm}
\begin{equation}
\label{eq:avg}
p^c = \sum_{k} w_k^c \cdot \sum_{i,j} y_{L}^{(k)}(i,j), \vspace{-.2cm}
\end{equation}
where $p^c$ is the probability output of class $c$. $(i,j)$ denotes spatial coordinates. $\bm w^c=[w_1^c, ..., w_k^c,...]^T$ is the $c$-th column of the weight matrix of the fully connected layer applied onto $p^c$.  $y_{L}^{(k)}$ is the $k$-th feature map of the last residual block. 
By plugging Eq. (\ref{eq:res_for}) into Eq. (\ref{eq:avg})\footnote{We omit the spatial coordinate $(i,j)$ and feature map dimension changes from $y_1$ to $\mathcal{F}_L$ for brevity.},  we can see that $p^c$ is the weighted average of the summed ensemble output:
\vspace{-.2cm}
\begin{equation}
\label{eq:avg_ens}
p^c = \sum_{i,j}  \bm w^c  y_{L} =  \sum_{i,j} \bm w^c  ( y_1+ \sum_{m=1}^{L-1} \mathcal{F}_{m} ). \vspace{-.2cm}
\end{equation}
In this paper, we argue that using a single weighting function in the classification module is suboptimal in this situation. This is because the outputs of all ensembles share classifiers such that the importance of their individual features are undermined. To address this issue, we propose to decouple the ensemble outputs and apply classifiers to them individually by using
\vspace{-.15cm}
\begin{equation}
p^c =  \sum_{i,j} \Big(\bm w^c_{1} \cdot  y_{1} + \sum_{m=1}^{L-1}  \bm w^c_{m+1} \cdot \mathcal{F}_{m} \Big). \vspace{-.15cm}
\end{equation}
Compared with Eq. (\ref{eq:avg_ens}), this equation assigns individual weight $\bm w^c_1$ to $\bm w^c_L$ for each ensemble output, which enables the classification module to independently decide the information importance from different residual blocks. 


We propose a "redesign'' of the ResNet architecture to realize the above idea, i.e., a new way to skip-connect a residual block, defined as follows: 
\vspace{-.1cm}
\begin{equation}
\label{eq:concat}
y_{l+1} = \mathcal{F}_{l}(y_{l}) \otimes y_{l}, \vspace{-.1cm}
\end{equation}
where $\otimes$ is the concatenation operation. We define this skip-connection scheme as \textit{ensemble-connection}. It allows outputs from residual blocks to flow through concatenated feature maps directly to the classification layer in parallel (see Figure \ref{fig:archs}), such that the classification module assigns weights to all network ensemble outputs and map them to the label space. It is straightforward to see that our design also ensures unimpeded information flow \cite{he2016identity} to overcome the gradient vanishing effect. 


We apply \textit{ensemble-connection} between residual blocks connecting block groups where the feature map dimension changes (see Appendix A) and maintain the identity mapping for blocks inside a group\footnote{Later on, we notice a new network, DenseNet \cite{huang2016densely}, which ends up with an analogous solution (concatenation replacing addition). We argue that our solution is based on a different motivation and results in a different architecture. Nevertheless, this network can be viewed as a successfully validation of our ensemble analysis.}. \textit{ensemble-connection} in nature integrates multi-scale representations in the last convolution layer. This multi-scaling scheme is essentially different from the skip output schemes used by \cite{xie2015holistically,bell2015inside}. 

\section{Language modeling and network training}
\vspace{-.1cm}
\subsection{Language model}
\label{sec:lang}
For language modeling, we use LSTM \cite{hochreiter1997long} to model the diagnostic reports by maximizing the joint probability over sentences:
\vspace{-.2cm}
\begin{equation}
\log p(\bm x_{0:T}|I;\theta_L)= \sum_{t=0}^{T}  \log p(\bm x_t|I, \bm x_{0:t-1}; \theta_L), \vspace{-.2cm}
\end{equation}
where $\{\bm  x_0,..., \bm  x_T\}$ are sentence words (encoded as one-hot vectors). The LSTM parameters $\theta_L$ are used to compute several LSTM internal states \cite{hochreiter1997long,vinyals2015show}. According to \cite{xu2015show}, we integrate the ``soft'' attention mechanism into LSTM through a context vector $\bm z_t$ (defined as follows) to capture localized visual information. To make prediction, LSTM takes the output of last time step $\bm x_{t-1}$ along with hidden state $\bm h_{t-1}$ and $\bm z_t$ as inputs, and computes the probability of next word $\bm x_t$ as follows:
\vspace{-.2cm}
\begin{equation} \vspace{-.2cm}
\begin{split}
&\bm h_t = LSTM(E \bm x_{t-1}, \bm h_{t-1}, \bm z_t), \\
 & p(\bm x_t|I,\bm x_{0:t-1}; \theta_L) \propto \exp(G_h \bm h_t), 
\end{split} 
\end{equation}
where $E$ is the word embedding matrix. $G_h$ decodes $\bm h_t$ to the output space. 

The attention mechanism dynamically computes a weight vector to extract partial image features supporting the word prediction, which is interpreted as an attention map indicting where networks capture visual information. Attention is the main component supporting the visual interpretability of our network. In practice, we observe that the original attention mechanism \cite{xu2015show} is more difficult to train, which often generates attention maps that smoothly highlight the majority of image area. 

To address this issue, we propose an auxiliary attention sharpening (AAS) module to improve its learning effectiveness. The attention mechanism can be viewed as a type of alignment between image space and language space. As indicted by \cite{liu2016attention}, improving such alignment can be achieved by adding supervision on attention maps by using region-level labels (e.g. bounding boxes). In order to deal with datasets that do not have any region-level labels, a new method needs to be developed. In our approach, rather than putting direct supervision on the weight vector $\bm a_t$, we propose to tackle this problem by utilizing the implicit class-specific localization property of global average pooling \cite{zhou2015learning} to support image-language alignment. Overall, $\bm z_t$ can be computed as follows:
\vspace{-.1cm}
\begin{equation} \vspace{-.1cm}
\label{eq:attention}
\begin{split}
\bm a_t  = softmax(& W_{att} \, \tanh (W_h \, \bm h_{t-1} + \bm c)),  \\
\bm c = & \; (\bm w^c)^T \mathcal{C}(I), \\
\bm z_t  = & \; \bm a_t \, \mathcal{C}(I)^T, \vspace{-.4cm}
\end{split}
\end{equation}
where $W_{att}$ and $W_h$ are learned embedding matrices. $\mathcal{C}(I)$ denotes Conv feature maps with dimension $512{\times}(14{\cdot}14)$ generated by the image model. $\bm c$ denotes a $196$-dimensional Conv feature embedding through $\bm w^c$.

\begin{figure}[t]
	\centering
	\includegraphics[width=0.48\textwidth]{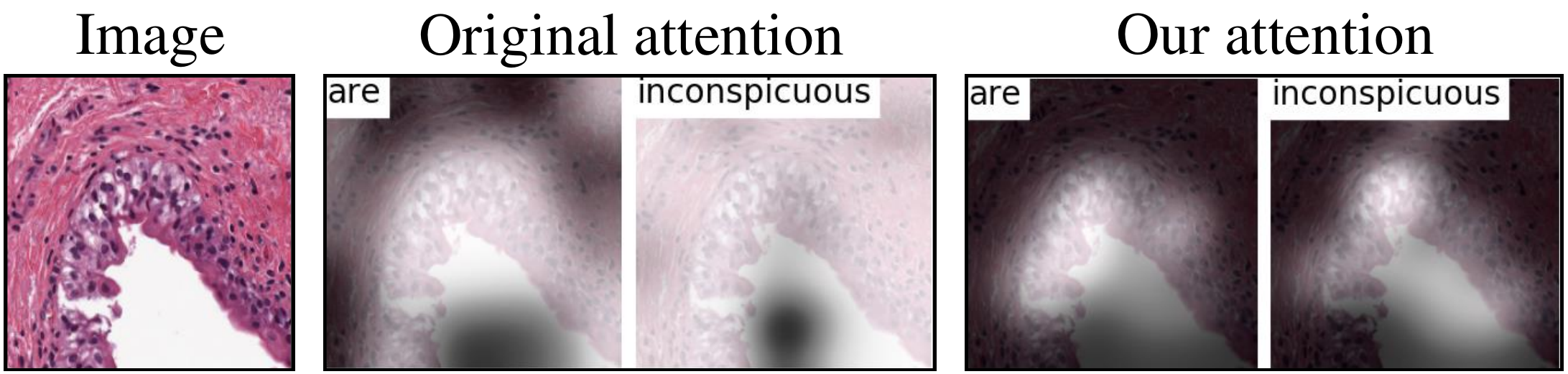}
	\vspace{-.7cm}
	\caption{The attention maps of the original method (middle) and our method (right).  Our method generates more focal attention on informative (urothelial) regions.  \vspace{-.5cm}} \label{fig:att_compare}
\end{figure}

The original attention mechanism learns $\bm w^c$  inside LSTM implicitly. In contract, AAS adds an extra supervision (defined in Section 4.2) to explicitly learn to provide more effective attention model training. Specifically, the formulation of this supervision is a revisit of Eq. (\ref{eq:avg_ens}) ($\mathcal{C}(I)$ stands for $y_L$; we use different notations for consistence).
$\bm w^c$ is a $512$-dimensional vector corresponding to the $c$-th column of the fully connected weight matrix, selected by assigned class $c$ (see Figure \ref{fig:archs}); when applied to $\mathcal{C}(I)$, the obtained $\bm c$ that carries class-specific and localized region information is used to learn the alignment with $\bm h_{t-1}$ and compute a $(14{\times}14)$-dimensional $\bm a_t$ and a $512$-dimensional context vector $\bm z_t$. 
 Figure \ref{fig:att_compare} compares the qualitative results between the original method and our proposed method.
\subsection{Effective gradient flow}
\label{sec:lstm-train}
\vspace{-.1cm}
In the well-known image captioning scheme \cite{karpathy2015deep,densecap}, a CNN provides an encoded image feature $F(I)$ as the LSTM input $\bm x_0$. Then a special START token is used as $\bm x_1$ to inform the start of prediction. Generating effective gradients w.r.t $F(I)$ is the key for the image model optimization.

A complete medical diagnostic report describes multiple symptoms of observing images, followed by the diagnostic conclusion about either one or multiple type of diseases. For example, radiology images have multiple disease labels \cite{shin2016learning}. Each symptom description specifically describes one type of image (symptom) feature. Effectively utilizing the semantic information in different descriptions is critical to generate effective gradient w.r.t $F(I)$ by LSTM.  

In our method, we let one LSTM focus on mining discriminative information from a specific description. All description modeling shares LSTM.
In this way, the modeling of each image feature description becomes a function of the complete report generation. We denote the number of functions as $K$. In the training stage,
given a mini-batch with $B$ pairs of image and reports, 
after forwarding the mini-batch to the image model, we duplicate each sample inside, resulting in a $K{\times}B$ mini-batch as the input of LSTM. Each duplication takes shared image features and one of $K$ types particular feature description  extracted from the report (see Figure \ref{fig:archs}).
The LSTM inputs of $\bm x^e_0$ and $\bm x^e_1$ are defined as
\vspace{-.1cm}
\begin{equation}
\bm  x_0^e  = W_F F(I), \; \; \bm  x_1^e = E S(e), 
\vspace{-.1cm}
\end{equation}
where $W_F$ is a learned image feature embedding matrix. $S(e), \,e=\{1,..., K\}$ is the one-hot representation of the $e$-th image feature type. In this way, we use particular $\bm x_1^e$ to inform LSTM the start of a targeting task. During backpropagation, the gradients w.r.t $F(I)$ from duplications are merged. All the operations are end-to-end trainable. 

To train AAS, we use the diagnostic conclusion as labels. The motivation are two-fold. First, the Conv feature embedding generated by AAS is specific to conclusion labels. Since all symptom descriptions support the inference of conclusion labels, it in nature contains necessary visual information to support different types of symptom descriptions and thereby can facilitate better alignment with description words in the attention model. Second, AAS serves as an extra supervision on the image model, which makes sure the image model training towards to optimal diagnostic conclusion.

\subsection{Network optimization}
\label{sec:optim}
The overall model has three sets of parameters: $\theta_D$ in the image model $D$, $\theta_L$ in the language model $L$, and $\theta_M$ in the AAS module $M$.  The overall optimization problem in MDNet is defined as 
\begin{equation}
\begin{split}
\max_{\theta_L, \theta_D, \theta_M} \; \mathcal{L}_{M}(l_c, M(D(I;\theta_D);\theta_M)) \\
+ \, \mathcal{L}_{L}(l_s, L(D(I;\theta_D);\theta_L)),
\end{split}
\end{equation}
where $\{I, l_c, l_s\}$ is a training tuple: input image $I$, label $l_c$ and groundtruth report sentence $l_s$. Modules $M$ and $L$ are supervised by two negative log-likelihood losses $\mathcal{L}_{M}$ and $\mathcal{L}_{L}$, respectively. 

The updating processes of $\theta_M$ and $\theta_L$ are independent and straightforward using gradient descent. Updating $\theta_D$ involves the gradients from both modules. We develop a backpropagation scheme to allow their composite gradients co-adapted mutually. Compared with \cite{ganin2014unsupervised}, the gradients in our method is calculated based on a mixture of a recurrent generative network and a multilayer perceptron.  
Specifically, $\theta_D$ is updated as follows: 
\begin{equation}
\label{eq:comp}
\theta_D \leftarrow \theta_D - \lambda \cdot \Big((1-\beta) \cdot \frac{\partial \mathcal{L}_M}{\partial \theta_D} + \beta \cdot \eta  \frac{\partial \mathcal{L}_L}{\partial \theta_D}\Big), 
\end{equation}
where $\lambda$ is the learning rate, and $\beta$ dynamically regulates two gradients during the training process. We also introduce another factor $\eta$ to control the scale of $\frac{\partial \mathcal{L}_L}{\partial \theta_D}$, because $\frac{\partial \mathcal{L}_L}{\partial \theta_D}$ often has smaller magnitude than $\frac{\partial \mathcal{L}_M}{\partial \theta_D}$. We will analyze the detailed configuration of these two hyperparameters and demonstrate the advantages of our proposed strategy.

\section{Experimental Results}
\vspace{-.1cm}
In this section, we start by validating the proposed image model (denoted as EcNet and explained in Section \ref{sec:ecnet}) of MDNet on the two CIFAR datasets that is specific for image recognition, with the purpose to show its superior performance against several other CNNs.
Then, we conduct sufficient experiments to validate the proposed full MDNet for medical image and diagnostic report modeling on the BCIDR dataset. Our implementation is based on Torch7 \cite{collobert2011torch7}. Please refer to Appendix for complete details.

\subsection{Image recognition on CIFAR} 
\vspace{-.1cm}
\label{sec:cifar}
We use well-known CIFAR-10 and CIFAR-100 \cite{krizhevsky2009learning} to validate our proposed EcNet. We follow the common way \cite{he2016identity} to process data and adopt the learning policy suggested by wide-ResNet (WRN) \cite{zagoruyko2016wide}. To choose baseline ResNet architectures, we consider depth as well as width to trade-off the memory usage and training efficiency \cite{zagoruyko2016wide}. We adopt the bottleneck residual block design instead of the ``tubby''-like block with two $3{\times}3$ convolution layers used by WRN, since we observe the former offers consistent improvement. We hypothesize that it is because the bottleneck design compacts information of feature maps doubled by \textit{ensemble-connection} (due to its concatenation operation), which promotes more efficient feature usage. Detailed architecture illustration is provided in Appendix A.


\begin{figure}[t]
	\begin{center}
		\includegraphics[width=0.40\textwidth,height=0.24\textwidth]{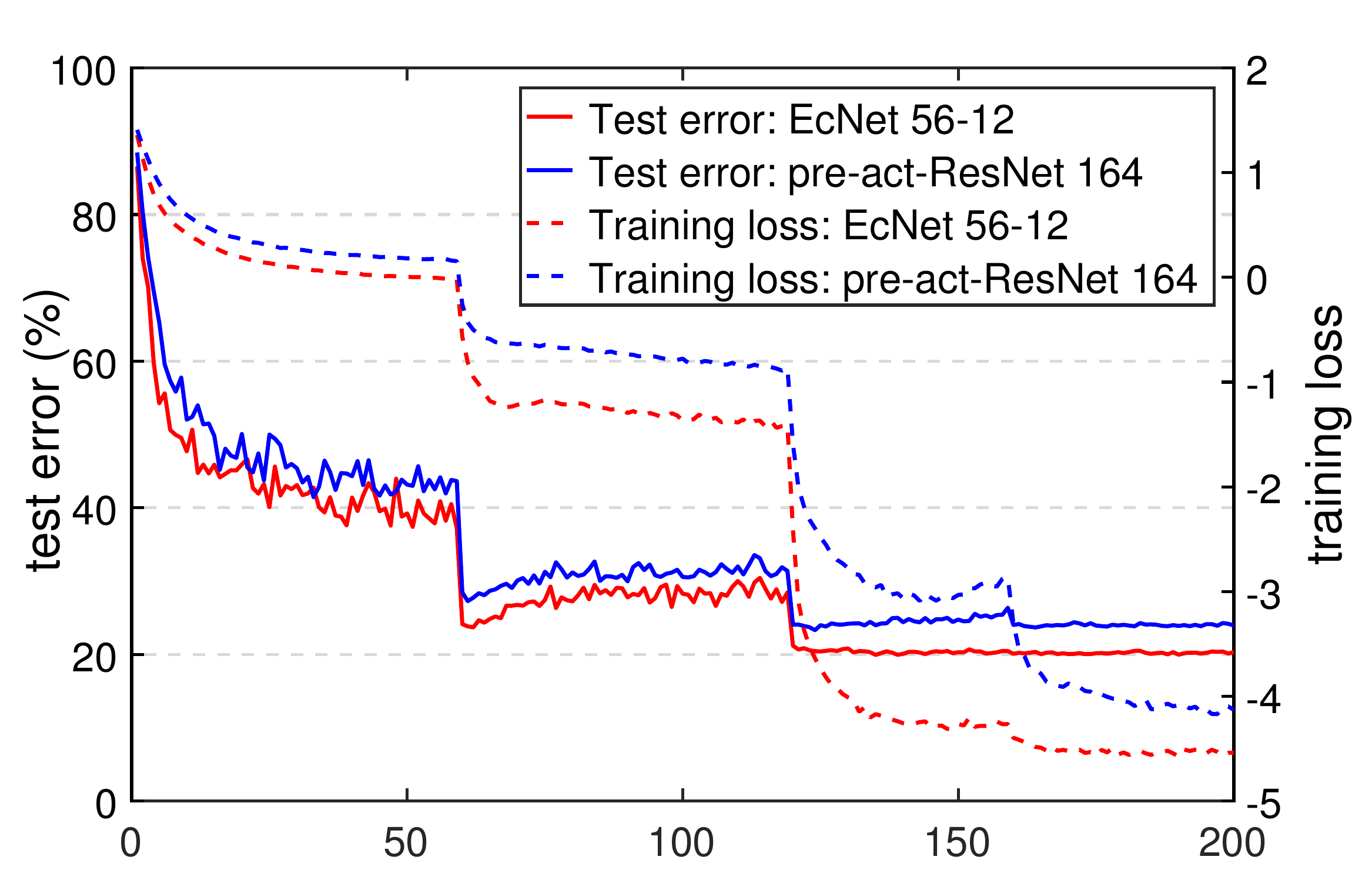}
	\end{center}
	\vspace{-.7cm}
	\caption{Training curves for CIFAR-100. \vspace{-.3cm} } \label{fig:curves}
\end{figure}
\begin{table}[t]
	\begin{center}
		\begin{tabularx}{.43\textwidth}{l|cccc}
			\specialrule{1.5pt}{0pt}{0pt}
			Method                                                     &  D-W  & Params &     C-10      &     C-100      \\
			\specialrule{1.0pt}{0pt}{0pt}
			NIN \cite{lin2013network} 								   &   -   &   -    &     8.81      &     35.67      \\
			Highway \cite{srivastava2015highway}                       &   -   &   -    &     7.72      &     32.39      \\ \hline
			ResNet  \cite{he2015deep}                                  &  110  &  1.7M  &     6.43      &     25.16      \\
			
			ResNet$^+$    \cite{he2016identity}                    &  164  &  1.7M  &     5.46      &       24.33        \\
			ResNet$^+$        \cite{he2016identity}                & 1001  & 10.2M  &     4.92      &     22.71      \\ \hline
			WRN     \cite{zagoruyko2016wide}                           & 40-4  &  8.7M  &     4.53      &     21.18      \\ \hline
			EcNet                                                       & 110-4 &  1.8M  &     4.91      &     22.53      \\
			EcNet                                                       & 56-12 &  8.0M  & \textbf{4.43} & \textbf{19.94} \\ \hline
		\end{tabularx}
		
	\end{center}
	\vspace{-.6cm}
	\caption{The error rate ($\%$) on CIFAR-10 (C-10) and CIFAR-100 (C-100). ResNet$^+$ denotes pre-act-ResNet.  The second column indicates network Depth-Width. Our result is tested on one trial.  \vspace{-.5cm} }\label{table:error}
\end{table}

Since this experiment is not the main focus of this paper, we left full architecture exploration for future work. We present two variants having similar number of parameters with compared variants of ResNet and WRN. The first one has depth $110$ and width $4$ and the second has depth $56$ and width $12$. Table \ref{table:error} compares the error rate on two datasets and Figure \ref{fig:curves} compares the training curves. Our EcNet-$56$-$12$ achieves obviously better error rate ($4.43\%$ in CIFAR-10 and $19.94\%$ in CIFAR-100) with only $8$M parameters compared with WRN-$40$-$4$ with $8.7$M parameters or ResNet$^{+}$-$1001$ with $10.2$M parameters. The results demonstrate that our \textit{ensemble-connection}, which enables the classification module to assign independent weights to network ensemble outputs, substantially
improves network ensembling effectiveness and, consequently, leads to higher efficiency of feature and parameter utilization. As mentioned in Section 1, these properties are favorable to medical images.

\begin{figure*}[t]
	\vspace{-.3cm}
	
	\begin {minipage}{0.475\textwidth}
	\vspace{-.0cm}
	\centering
	\includegraphics[width=1\textwidth, height=.74\textwidth]{attention4.pdf}
	\vspace{-.6cm}
	\caption{The image model predicts diagnostic reports (left-up corner) associated with sentence-guided attention maps. The language model attends to specific regions per predicted word. The attention is most sharp on urothelial neoplasms, which are used to diagnose the type of carcinoma.
	}\label{fig:attention2}
\end{minipage}\hfill
\begin{minipage}{0.48\textwidth}
	\centering
	\includegraphics[width=1\textwidth,height=0.7\textwidth]{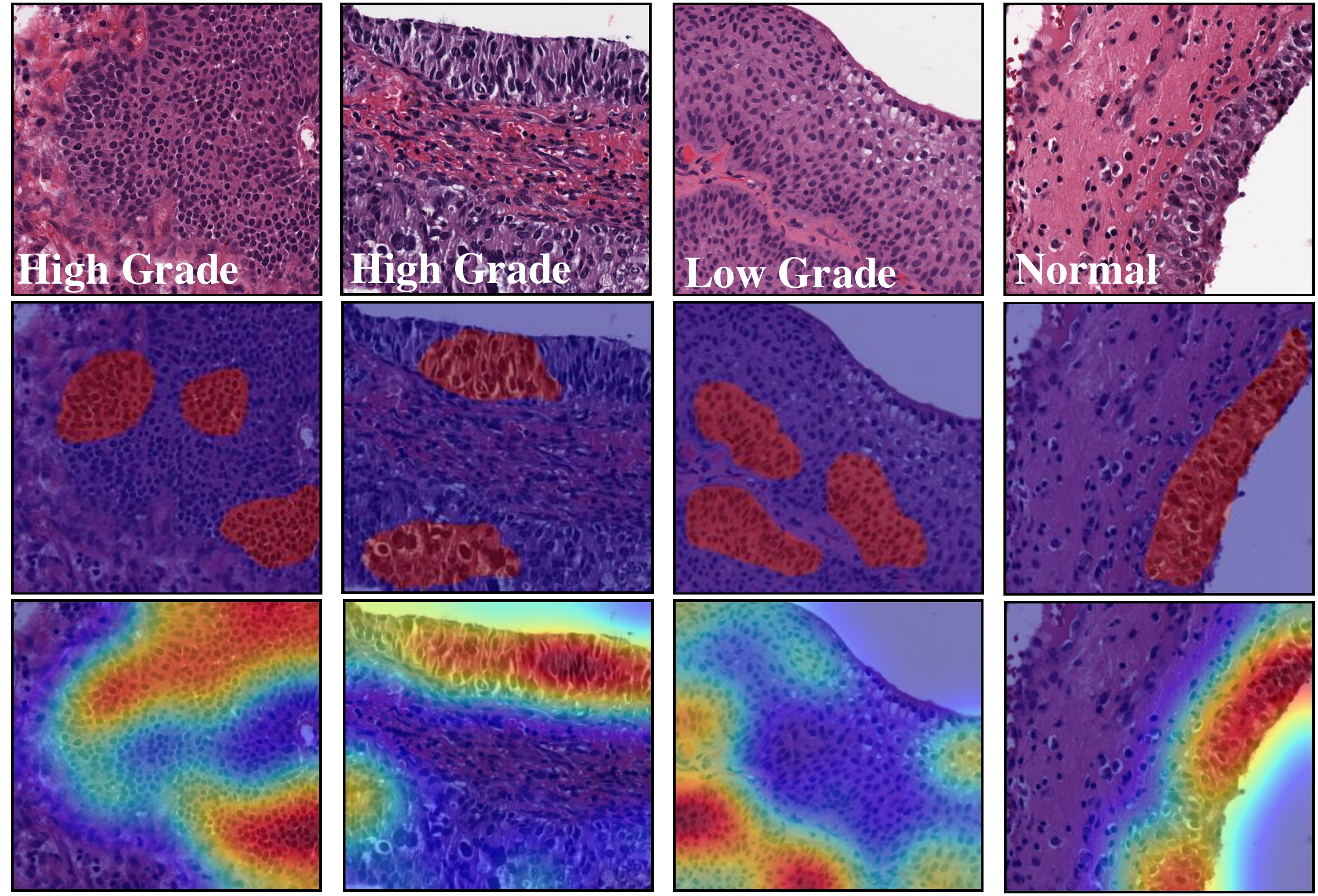}
	\vspace{-.6cm}
	\caption{The illustration of class-specific attention. From top to bottom, test images, pathologist annotations, and class attention maps. Like the pathologist annotations, the attention maps are most activated in urothelial regions, largely ignoring stromal or background regions. Best viewed in color. }\label{fig:attention1}
\end{minipage}
\vspace{-.5cm}
\end{figure*}

\subsection{Image-language evaluation on BCIDR}
\label{sec:dataset}
\vspace{-.15cm}
We evaluate our MDNet for two tasks: report generation and symptom based image retrieval. We follow common evaluation methods \cite{mao2014deep} but also suggest complementary evaluation metrics specially designed for medical images. To validate our method, we use $5$-fold cross validation. Appendix B discusses training details.

\noindent
\textbf{Dataset} The bladder cancer image and diagnostic report (BCIDR) dataset was collected in collaboration with a pathologist. Whole-slide images were taken using a 20X objective from hematoxylin and eosin (H$\&$E) stained sections of bladder tissue extracted from a cohort of 32 patients at risk of a papillary urothelial neoplasm. From these slides, 1000 500x500 RGB images were randomly extracted close to urothelial neoplasms (each slide yields a slightly different number of images). 
We used a web interface to show each image (without diagnostic information of patient slides) and the pathologist then provided a paragraph describing observations to address five types of cell appearance features (Figure \ref{fig:archs} shows an example), namely the state of nuclear pleomorphism, cell crowding, cell polarity, mitosis, and prominence of nucleoli followed by a diagnostic conclusion. The conclusion is comprised of four classes, i.e., normal, papillary urothelial neoplasm of low malignant potential (PUNLMP)/low-grade carcinoma, high-grade carcinoma, and insufficient information. Following this procedure, four doctors (non-experts in bladder cancer) wrote an additional four descriptions in their own free words but referring to the pathologist's description to guarantee accuracy. Thus there are five ground-truth reports per image in total. Each report varies in length between 30 and 59 words. 

We randomly select $20\%$ (6/32) of patients including $200$ images as testing data and the remaining $80\%$ of patients including $800$ images for training and cross-validation. For data processing, the input image is resized to $224{\times}224$. We subtract the RGB mean from each image and augment the training data through clip, mirror and rotation operations. According to this dataset, the five descriptions and one conclusion are treated as $K{=}6$ separate tasks (defined in Section \ref{sec:lstm-train}) for LSTM training to support complete report generation. The conclusion is used as (4-way) labels for CNN training in all comparison experiments.

\noindent
\textbf{Baseline} We choose the well-known image captioning scheme \cite{karpathy2015deep,vinyals2015show} (the source code of \cite{karpathy2015deep}) as the baseline, which is to first train a CNN to represent images, followed by training an LSTM to generate descriptions. We use GoogLeNet instead of its originally used VGG \cite{simonyan2014very}, since the former performs better on BCIDR. We also train a small version of our EcNet, which has depth $38$ and width $8$, including $2.3$M parameters (our purpose here is not to compare EcNet and GoogLeNet). Pre-trained GoogleNet and EcNet per validation fold are shared by all comparative models. When training LSTM, we test the cases with and without fine-tuning CNNs. 

\noindent
\textbf{Ablation study} MDNet is jointly trained which needs no pre-training or fine-tunning. For detailed comparison with the baseline, we also test two cases which training MDNet using the baseline strategies. In these cases, our optimization is not applied, so the differences from the baseline are task-separated LSTM and the integrated attention model. 

\begin{table*}[ht]
	\begin{center}
		
		\begin{tabularx}{.838\textwidth}{c|cccc|ccccccc|c}
			\specialrule{1.5pt}{0pt}{0pt}
			Model & CNN &     P?     &     F?  & J?   &      B1 &B2 &B3 &B4     &       M       &       R       &       C       &  DCA($\%$)$\pm$std   \\ \hline
			\multirow{4}{*}{Baseline}                   
			&  GN  & \checkmark &         &   &     90.6&81.8&73.9&66.6      &     39.3     &     69.5      &     2.05      &     72.6$\pm$1.8      \\
			&  GN  & \checkmark & \checkmark &  &   90.7&82.0&74.3&66.9   &    39.5      &     69.9      & \textbf{2.09} &     74.2$\pm $3.8      \\
			&    EN    & \checkmark &         &   			&     90.1&81.1&73.2&65.8      &     39.3      &     69.7      &     2.01     &     73.7$\pm$2.4     \\
			&    EN    & \checkmark & \checkmark & &    90.3&81.9&74.1& 66.8      &     39.6      &     69.8      &     2.02      &     74.4$\pm$4.8     \\ \hline\hline
			\multirow{3}{*}{Ours}                     
			&    EN    & \checkmark &        &    &     90.4&81.9&74.1&66.6      &     39.3     &     69.8      &     1.95      &     72.7$\pm$4.2      \\
			&    EN    & \checkmark & \checkmark &  &   90.4&81.5&73.4&65.9      &     39.0      &     69.5      &     1.92      &     71.6$\pm$4.2      \\
			&    EN    &            &            & \checkmark &\textbf{91.2}&\textbf{82.9 }&\textbf{75.0}&\textbf{67.7}&\textbf{39.6} & \textbf{70.1} &     2.04      &    \textbf{ 78.4}$\pm$1.5     \\ \hline
		\end{tabularx}
	\end{center}
	\vspace{-.6cm}
	\caption{Quantitative evaluation of generated description quality and the DCA score. See text for metric notations. P, F, and J denote whether a pre-trained CNN is used, whether fine-tuning pre-trained CNNs when training LSTM, and whether using our proposed joint training approach (i.e. our proposed MDNet), respectively. The 5th and 6th rows are for the ablation study. GN and EN denote GoolgeNet and EcNet.  } \label{table:evaluation} \vspace{-.5em}
	\vspace{-.2cm}
\end{table*}
\begin{table}[t] 
	\begin{center}
		\newcolumntype{C}{>{\centering\arraybackslash}p{8ex}}
		\newcolumntype{T}{>{\centering\arraybackslash}p{3.5ex}}
		\begin{tabularx}{.487\textwidth}{p{1ex}|cp{1ex}p{1ex}p{1ex}|CCC}
			\specialrule{1.5pt}{0pt}{0pt}            
			& CNN  & P?    & F?  & J?   &     Cr@1      &      Cr@5      &     Cr@10      \\ \hline
			\multirow{4}{*}{\rotatebox[origin=c]{90}{Baseline}} & GN & \checkmark &       &     &     71.7$\pm2.5$      &      71.9$\pm5.2$       &      72.9$\pm4.1$       \\
			& GN & \checkmark & \checkmark & &     70.1$\pm8.3$     &      72.5$\pm5.9$      &      72.8$\pm5.3$      \\
			& EN     & \checkmark &        			&    &     64.4$\pm2.4$     &      70.8$\pm0.9$      &      72.5$\pm1.6$     \\
			
			& EN     & \checkmark & \checkmark &   &  68.3$\pm2.0$      &      71.8$\pm1.5$      &      73.4$\pm1.9$      \\ \hline\hline
			\multirow{3}{*}{\rotatebox[origin=c]{90}{Ours}}     
			& EN     & \checkmark &   		  & &       68.7$\pm5.5$           &       73.1$\pm2.8$         &  74.3$\pm1.7$ \\
			& EN     & \checkmark & \checkmark & &    71.6$\pm5.5$         &    75.7$\pm3.9$            &       75.8$\pm2.7$         \\
			& EN    &            &        &   \checkmark  & \textbf{78.6}$\pm4.0$ & \textbf{79.5}$\pm3.6$  & \textbf{79.4}$\pm3.1$  \\ \hline
		\end{tabularx}
	\end{center}
	\vspace{-.5cm}
	\caption{Quantitative evaluation (mean$\pm$std) of report to image retrieval. See text for explanation of the metric Cr@$k$. The last row is our proposed MDNet.  \vspace{-.5cm}} \label{table:retrieval} 
\end{table}

%
\vspace{-.5cm}
\subsubsection{Interpret model prediction}
\vspace{-.2cm}
We start by qualitatively demonstrating the diagnosis process of MDNet: generating reports and showing image attention to interpret how the network uses visual information to support its diagnostic prediction. Two kinds of attention maps are demonstrated. 

Sentence-guided attention is computed by our attention model, where each attention map corresponds to a predicted word to show the relevant part of image that the network attend. According to pathologists' observations, our computed attention maps are fairly encouraging, which intend to attend on informative regions and avoid less useful regions. Figure \ref{fig:attention2} shows sample results. Please see the supplementary material for more results.

The conclusion-specific attention map is computed by AAS (i.e. the $14{\times14}$ Conv feature embedding). Recall that it has the implicit localization ability on image parts relate to the predicted label. To evaluate this attention qualitatively, we ask the pathologist to draw regions of interest of some test images that is necessary to infer conclusion based on his experience. Figure \ref{fig:attention1} shows the results. There is fairly strong correspondence between the pathologist annotations and regions with the sharpest attention. Recall that the training stage does not have region level annotations. These results demonstrate that MDNet has learned to discover useful information to support its prediction.

\vspace{-.5cm}
\subsubsection{Diagnostic report generation}
\label{sec:report}
\vspace{-.2cm}
\noindent
\textbf{Evaluation metrics} We report commonly used image captioning evaluation metric scores \cite{vedantam2015cider}, including BLEU(B), METEOR(M), Rouge-L(R), and CIDEr(C). The diagnostic reports have more regular linguistic structure than natural image captions. Our experiments show that standard LSTM can capture the general structure, resulting in similar metric scores. Nevertheless, we care more about whether the trained models accurately express pathologically meaningful keywords. To make more definitive evaluation, we report the predicted diagnostic conclusion accuracy (DCA) extracted from generated report sentences. 

The results are shown in Table \ref{table:evaluation}. Our proposed MDNet (last row) outperforms all comparative baseline models by demonstrating significantly improved DCA (also smaller std) and most of other metrics. For the baseline methods in the first block of the table, the models using EcNet (3th and 4th rows) achieve slightly better results than the models using GoogLeNet. We also observe that fine-tuning the pre-trained CNNs (either EcNet and GoogleNet) is generally beneficial but more unstable (i.e. higher std). The following image retrieval experiments provide more quantitative evaluation of the sentence-image mapping quality.

\begin{figure*}[t]	
	\begin{center}
		
		\begin{subfigure}[b]{0.31\textwidth}
			\centering
			\includegraphics[width=0.99\textwidth,height=0.62\textwidth]{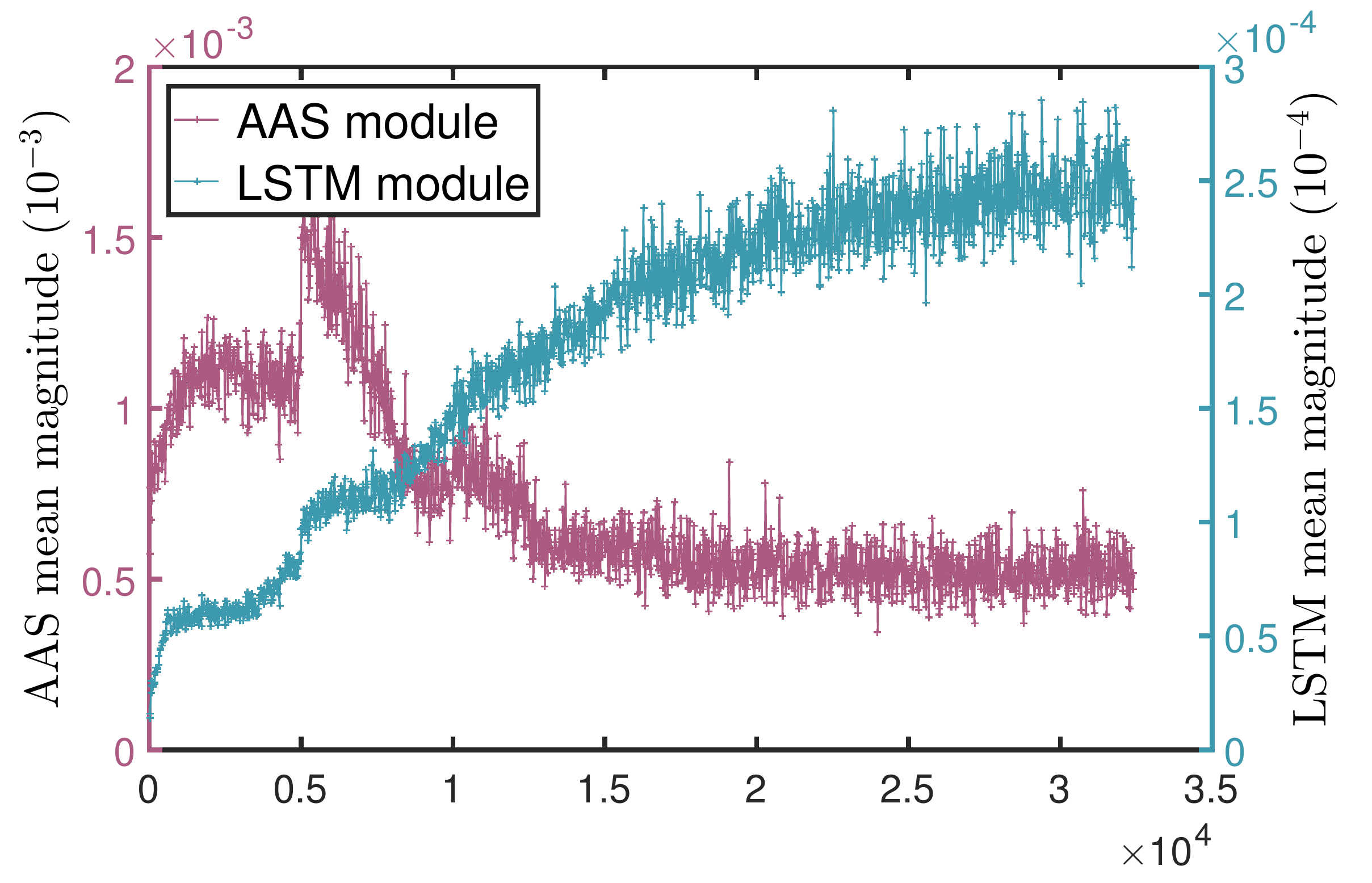}
		\end{subfigure}
		\begin{subfigure}[b]{0.31\textwidth}
			\centering
			\includegraphics[width=.999\textwidth,height=0.60\textwidth]{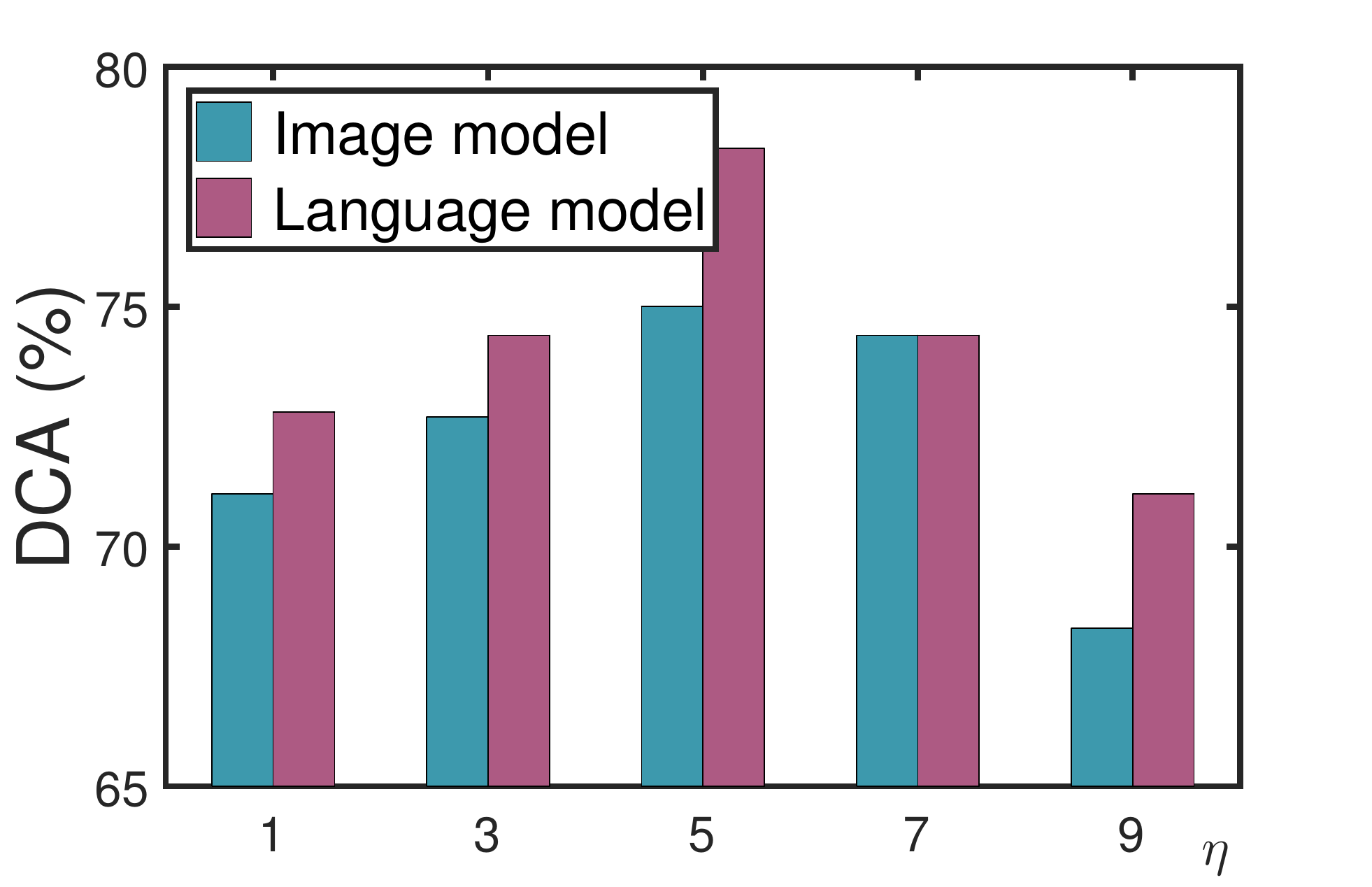}
		\end{subfigure}
		\begin{subfigure}[b]{0.31\textwidth}
			\centering
			\includegraphics[width=.999\textwidth,height=0.60\textwidth]{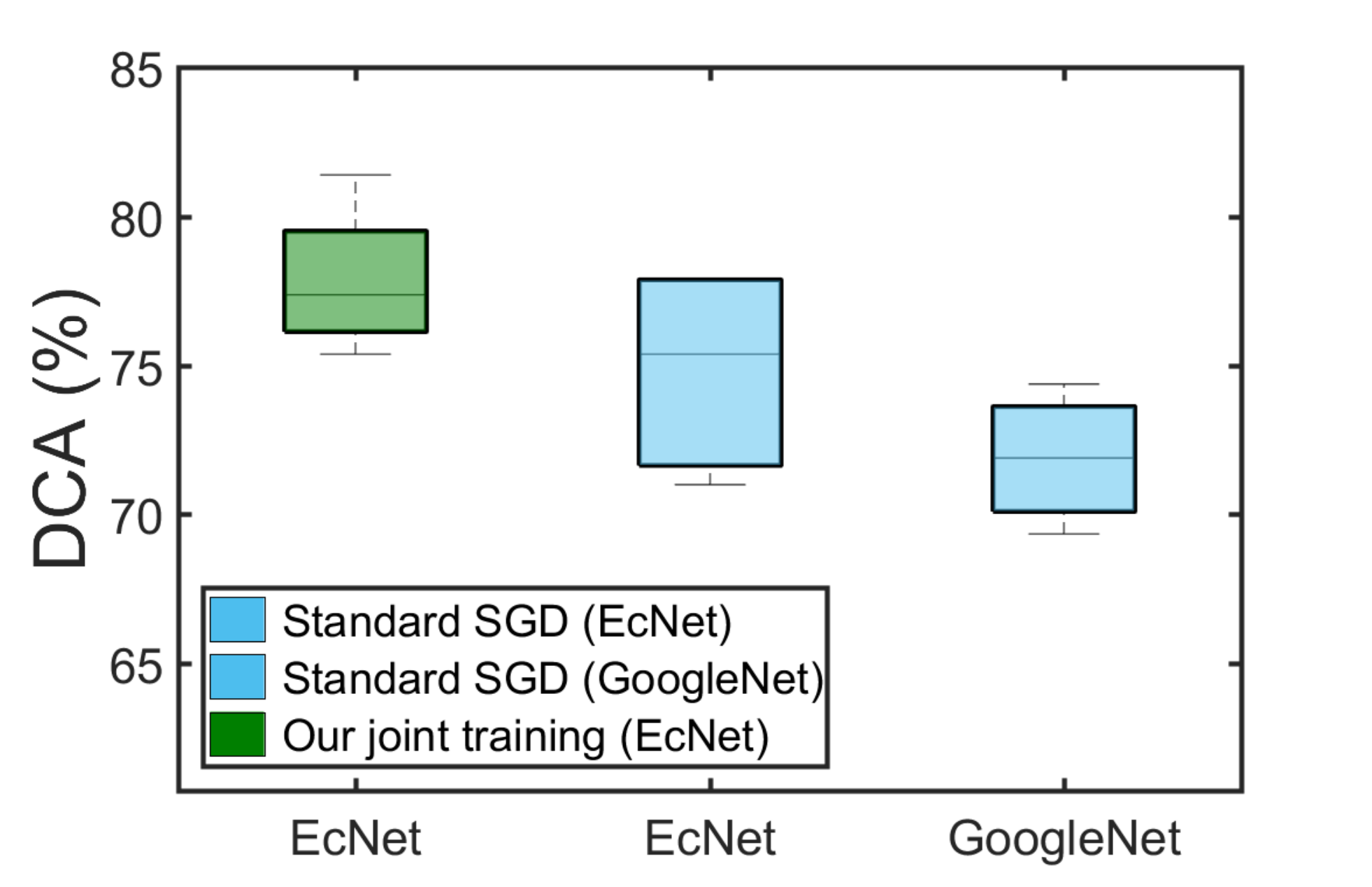}
		\end{subfigure}
		
	\end{center}
	\vspace{-.6cm}
	\caption{\textbf{Left}: The mean gradient magnitude. \textbf{Middle}: The DCA scores of the image model and language model in a MDNet respect to different $\eta$ in x-axis. \textbf{Right}: The DCA (over 5 folds) scores of EcNet (stands for the image model of MDNet) and pre-trained EcNet and GoogLeNet. \vspace{-.4cm}} \label{fig:magnitude}
\end{figure*}

\vspace{-.5cm}
\subsubsection{Symptom description based image retrieval}
\vspace{-.2cm}
We evaluate all trained models in Table \ref{table:evaluation} for the symptom description based image retrieval task shown in Table \ref{table:retrieval}.

\noindent
\textbf{Evaluation metric} Natural image captioning methods evaluate the groundtruth image recall at top $k$ positions  based on the ranking of images given an query sentence \cite{hodosh2013framing,mao2014deep}.
However, in the medical image domain, this metric is not necessarily valid because images with close symptoms could share similar descriptions. Thus, low recall does not exactly indicate poor models. Instead, we evaluate the ability of the model to retrieve images with correct diagnostic conclusion given a query report. But for all query reports, we remove the words related to conclusion and only keep image feature descriptions. The intuition behind this metric is that doctors have clinical needs to query images with specified symptoms. Given some diseased image descriptions, it should be a failure if the model retrieves a healthy image. This metric is an exact measurement of sentence-image mapping quality because a mistake in a single symptom description could result in retrieval errors. We report the correct conclusion recall rate, denoted as Cr@$k$, $k =\{1,5,10\}$, of top $k$ retrieved images corresponding to the query report. 

Table \ref{table:retrieval} shows the mean (std) scores over 5 folds. As can be observed, fine-tuning EcNet results in noticeable improvement generally, especially for the two experimental cases on our network (5th and 6th rows), though they do not reach the results of our proposed MDNet (last row). 
Based on present results, we observe: 
\begin{enumerate}
	\vspace{-.1cm}
	\itemsep-.1cm
	\item In general, fine-tuning pre-trained EcNet gives rise to larger improvement than fine-tuning GoogLeNet. 
	\item MDNet that separates the modeling of overall reports as functions of independent image descriptions is more accurate to capture fine discrimination in descriptions, while fine-tuning ($6$th row against $5$th row) further improves the mapping quality thanks to the design in Section \ref{sec:lstm-train}.
	\item Our proposed MDNet significantly outperforms baseline models, which indicates much better sentence-image mapping quality. One reason is because our joint training method prevents overfitting effectively. 
	\vspace{-.4cm}
\end{enumerate}

\vspace{-.3cm}
\section{Discussion}
\label{sec:analysis}
\vspace{-.1cm}
 
\noindent
\textbf{Optimization} The weight of composite gradients are shifting during training. The basic rule is to assign large weight to $\frac{\partial \mathcal{L}_M}{\partial \theta_D}$ to allow AAS to dominate the image model training for a while, and gradually increase the scale of $\frac{\partial \mathcal{L}_L}{\partial \theta_D}$ to introduce semantic knowledge and facilitate two models co-adapt mutually. We use a sigmoid-like function to change $\beta$ from $0$ to $1$ gradually during the entire training process. 

Balancing the scale of the two gradients is critical. We observe that simply scaling up $\frac{\partial \mathcal{L}_L}{\partial \theta_D}$ without scaling down $\frac{\partial \mathcal{L}_M}{\partial \theta_D}$ (i.e. remove $1-\beta$) has negative effects in our practice, probably because the totally summed gradient w.r.t $\theta_D$ will grow larger and increase instability in model training. We observe ${\sim}4\%$ DCA score decrease  of the language model without averaging. Thus, we argue that using weighted averaging is necessary.
However, simply averaging two gradients (using $\beta$) will make $\frac{\partial \mathcal{L}_M}{\partial \theta_D}$ overwhelm $\frac{\partial \mathcal{L}_L}{\partial \theta_D}$ since they have different magnitudes. A heuristic way to observe this fact is to visualize their mean gradient magnitudes. As can be observed in Figure \ref{fig:magnitude}(left), the gradient magnitude of $\frac{\partial \mathcal{L}_L}{\partial \theta_D}$ is much smaller than that of $\frac{\partial \mathcal{L}_M}{\partial \theta_D}$. We cross-validated $\eta$ (see Figure \ref{fig:magnitude}(middle)) and set $\eta=5$ throughout. 

\noindent
\textbf{Small dataset and regularization} The size of BCIDR is much smaller than common natural image datasets. This situation yields higher possibilities to end up with overfitted models, though we use regularization techniques and cross-validation. However, small dataset size is a common issue in the medical image domain; these large networks are still widely used \cite{shin2016learning,shin2016deep}. Figuring out effective regularization is extremely necessary. Both pre-trained CNNs and the image model of MDNet (i.e. AAS outputs) predict diagnostic conclusion labels. We can utilize this definite DCA score for more detailed analysis and comparison. 

For all trained models, we observe the DCA of the language model strongly relies on that of corresponding image model (see Figure \ref{fig:magnitude}(middle)), which motivates us to analyze more about CNN training itself. According to Eq. (\ref{eq:comp}), module $M$ provides a standard CNN loss. If we interpret $\frac{\partial \mathcal{L}_L}{\partial \theta_D}$ from module $L$ as ``noise'' added onto the gradient $\frac{\partial \mathcal{L}_M}{\partial \theta_D}$, this ``noise'' disturbs the loss of module $M$ and overall CNN training. In fact, moderate disturbance on the loss layer has regularization effects \cite{xie2016disturblabel}. Therefore, our optimization behaves particular regularization on CNN to overcome overfitting. As compared in Figure \ref{fig:magnitude}(right), the image model of MDNet trained using our optimization approach outperforms pre-trained CNN models using stochastic gradient descent (SGD). 

\noindent
\textbf{Multimodal mapping for knowledge fusion} Image feature descriptions in diagnostic reports contain strong underlying supports for diagnostic conclusion inference. According to our results, our proposed MDNet for multimodal mapping learning effectively utilizes these semantic information to encourage sufficient multimodal knowledge sharing between image and language models, resulting in better mapping quality and more accurate prediction.

\vspace{-.1cm}
\section{Conclusion and Future Work}
\vspace{-.1cm}
This paper presents a novel unified network, namely MDNet, to establish the direct multimodal mapping from medical images and diagnostic reports. Our method provides a novel perspective to perform medical image diagnosis: generating diagnostic reports and corresponding network attention, making the network diagnosis and decision-making process semantically and visually interpretable. Sufficient experiments validate our proposed method.

Based on this work, limitations and open questions are drawn: building and testing large-scale pathology image-report datasets; generating finer \cite{simonyan2013deep} attention for small biomarker localization; applying to whole slide diagnosis. We expect to address them in the future work.

{\small
\bibliographystyle{ieee}
\bibliography{egbib}
}

\end{document}